\documentclass{tlp}

\def\bdm{\begin{displaymath}}
\def\edm{\end{displaymath}}
\def\satisfies{\models}

\def\ii#1{\hbox{\it #1\/}}
\def\no{\ii{not}}
\def\lar{\leftarrow}
\def\beq{\begin{equation}}
\def\eeq#1{\label{#1}\end{equation}}
\def\ba{\begin{array}}
\def\ea{\end{array}}

\newtheorem{thm}{Theorem}
\newtheorem{lem}{Lemma}
\newtheorem{prop}{Proposition}
\newtheorem{cor}{Corollary}

\begin{document}

\title[Tight Logic Programs]
{\bf Tight Logic Programs}
\author[Esra Erdem and Vladimir Lifschitz]
       {Esra Erdem \\
        Department of Computer Science \\
        University of Toronto \\
        6 King's College Road \\
        Toronto, ON M5S 3H5, Canada \\
        \email{esra@cs.toronto.edu}
        \and
        Vladimir Lifschitz\\
	Department of Computer Sciences \\
        University of Texas at Austin\\
        Austin, TX 78712, USA \\
        \email{vl@cs.utexas.edu}}

\date{}
\maketitle

\begin{abstract}
This note is about the relationship between two
theories of negation as failure---one based on program completion, the other
based on stable models, or answer sets.  Fran\c{c}ois Fages showed that if
a logic program satisfies a certain syntactic condition, which is now
called ``tightness,'' then its stable models can be characterized
as the models of its completion.  We extend the definition of tightness
and Fages' theorem to programs with nested
expressions in the bodies of rules, and study tight logic programs
containing the definition of the transitive closure of a predicate.
\end{abstract}

\section{Introduction} \label{sec:introduction}

This note is about the relationship between two
theories of negation as failure---one based on program completion
\cite{cla78}, the other based on stable models, or answer sets
\cite{gel88}.  Fran\c{c}ois Fages~(1994)
showed that if a logic program satisfies a certain syntactic
condition, which is now called ``tightness,''\footnote{The term used by
Fages is ``positive-order-consistent.''} then its stable models can be
characterized as the models of its completion.  Lifschitz~(1996)
observed that Fages' theorem can be extended to programs with
infinitely many rules and to programs with classical negation~\cite{gel91b}
if the concept of completion in the statement of the theorem is replaced by
its semantic counterpart---the concept of a supported model~\cite{apt88}.
Fages' theorem was further generalized in~\cite{bab00}.

In this paper we show how to extend Fages' theorem to programs with nested
expressions in the bodies of rules.  A generalization of the completion
semantics to such programs was proposed in~\cite{llo84a}, and a similar
generalization of the answer set semantics is given in~\cite{lif99d}.
Here is an example. Program
\beq
\ba l
p \lar \no\ \no\ p \\
p \lar p,q
\ea
\eeq{ex}
contains nested occurrences of negation as failure in the body of the
first rule.\footnote{The double negation in the first rule of (\ref{ex})
is redundant from the point of view of the completion semantics, but it
does affect the program's answer sets. On the other hand, the second rule
is redundant from the point of view of the answer set semantics.  But,
generally, dropping a rule like this can change a program's completion in
an essential way.} It belongs to the syntactic class for which our theorem
guarantees the equivalence of the answer set semantics to the completion
semantics. This program has two answer sets $\emptyset$, $\{p\}$; they are
identical to the models of the completion
\beq
\ba l
p \equiv \neg \neg p \vee (p \wedge q) \\
q \equiv \bot
\ea
\eeq{ex-comp}
of this program.

A preliminary report on the tightness of programs with nested expressions
is published in~\cite{erd01a}.

The second question studied here is the tightness of logic programs
containing the definition of the transitive closure of a predicate:
$$
%\beq
\ba l
\ii{tc}(x,y) \lar p(x,y)\\
\ii{tc}(x,y) \lar p(x,v), \ii{tc}(v,y).
\ea
%\eeq{trans-p}
$$
Such rules are found in many useful programs.  Unfortunately, the definition of
tightness may be difficult to verify directly for a program containing
these rules.  We give here a sufficient condition that can make this easier.  A
preliminary report on this part of the work was presented at the 2001 AAAI
Spring Symposium on Answer Set Programming~\cite{erd01}.

The concept of a tight program and generalizations of Fages' theorem may be
interesting for two reasons.

First, the completion semantics and the stable model semantics are among the
most widely used definitions of the meaning of negation as failure, and it
is useful to know under what conditions they are equivalent to each other.

Second, the class of tight programs is important from the perspective of
answer set programming.  Whenever the two semantics are equivalent, answer
sets for the program can be computed by solving the satisfiability problem
for its completion.  This can be done using a satisfiability solver such as
{\sc sato}\footnote{\tt http://www.cs.uiowa.edu/\~{}hzhang/sato.html .} or
{\sc mchaff}\footnote{\tt http://www.ee.princeton.edu/\~{}mchaff/ .}~\cite{bab00}.  The idea of computing answer sets for a program using a satisfiability solver
led to the creation of the system {\sc cmodels}\footnote{\tt
http://www.cs.utexas.edu/users/tag/cmodels.html .} which, in some cases,
finds answer sets faster than ``general-purpose'' answer set solvers
 such as
{\sc smodels}\footnote{\tt http://www.tcs.hut.fi/Software/smodels/ .} and
{\sc dlv}\footnote{\tt http://www.dbai.tuwien.ac.at/proj/dlv/ .}.
  Our generalization of Fages' theorem allows us to apply this idea to
programs containing weight constraint rules~\cite{sim02}, because such
programs can be viewed as a special case of programs with nested
expressions~\cite{fer03}.  For instance, rule $\{p\}$ can be treated as
shorthand for
$$p \lar \no\ \no\ p.$$
{\sc cmodels} uses this fact to
handle programs with weight constraints, and our generalization
of Fages' theorem allows {\sc cmodels} to decide whether the answer sets
for such a program are identical to the models of its completion.

We begin by reviewing the definitions of answer sets, closure, supportedness
and completion for programs with nested expressions
(Sections~\ref{sec:programs}, \ref{sec:supportedness}).  After discussing
the concept of tightness for such programs in Sections~\ref{sec:tightness}
and~\ref{sec:abs}, we
state our generalization of Fages' theorem (Section~\ref{sec:thm}).  As an
example, we
show how this generalized form applies to a formalization of the $n$-queens
problem (Section~\ref{sec:eightqueens}).
Then we study the tightness of programs containing the definition of the
transitive closure of a relation (Section~\ref{sec:tc}).  A logic programming
description of the blocks world is used as an example in Section~\ref{sec:bw}.
Proofs of theorems are given in Section~\ref{sec:proofs}.

\section{Programs and Answer Sets} \label{sec:programs}

This section is a review of the answer set semantics for nondisjunctive
programs with nested expressions.

The words {\sl atom} and {\sl literal} are understood here as in
propositional logic; we call the sign $\neg$ in a negative literal $\neg
A$ {\sl classical negation}, to distinguish it from the symbol for negation as
failure (\no). {\sl Elementary formulas} are literals and the 0-place
connectives $\bot$ and $\top$. {\sl Formulas} are built from elementary
formulas using the unary connective \no\ and the binary connectives ,
(conjunction) and ; (disjunction). A {\sl (nondisjunctive) rule} is an
expression of the form
\beq
\ii{Head} \lar \ii{Body}
\eeq{rule}
where $\ii{Head}$ is a literal or $\bot$, and $\ii{Body}$ is a
formula.\footnote{In \cite{lif99d}, the syntax of rules is more general:
the head may be an arbitrary formula, in particular a disjunction.} If
$\ii{Body}=\top$, we will drop both the body and the arrow separating it from
the head; rules with the body $\top$ are called {\sl facts}.
If $\ii{Head}=\bot$, we will drop the head; rules with the
head $\bot$ are called {\sl constraints}.

A {\sl (nondisjunctive logic) program} is a set of rules.

We define when a consistent set $X$ of literals {\sl satisfies} a formula
$F$ (symbolically, $X \satisfies F$) recursively, as follows:
\begin{itemize}
\item for elementary $F$, $X \satisfies F$ if $F \in X$ or $F  = \top$,
\item $X \satisfies \no\ F$ if $X \not\satisfies F$,
\item $X \satisfies (F,G)$ if $X \satisfies F$ and $X \satisfies G$,
\item $X \satisfies (F;G)$ if $X \satisfies F$ or $X \satisfies G$.
\end{itemize}
A consistent set $X$ of literals is {\sl closed under} a program~$\Pi$ if,
for every rule~(\ref{rule}) in $\Pi$, $\ii{Head} \in X$ whenever
$X \satisfies \ii{Body}$.

Let $\Pi$ be a program without negation as failure.  We say that
$X$ is an {\sl answer set} for $\Pi$ if $X$ is minimal among the
consistent sets of literals closed under $\Pi$. It is easy to see that there
can be at most one such set.  For instance, the answer set for the program
\beq
\ba l
p\\
p \lar p,q
\ea
\eeq{ex-r}
is $\{p\}$.

The {\sl reduct} $\Pi^X$ of a program $\Pi$ relative to a
set $X$ of literals is obtained from $\Pi$ by replacing every maximal
occurrence of a formula of the form $\no\ F$ in $\Pi$ (that is, every
occurrence of $\no\ F$ that is not in the scope of another \no)
with $\bot$ if $X \satisfies F$, and with $\top$ otherwise.\footnote{This
definition is equivalent to the recursive definition of the reduct given
in~\cite{lif99d}.}
A consistent set $X$ of literals is an {\sl answer set} for $\Pi$ if it is
the answer set for the reduct $\Pi^X$. For instance, $\{p\}$ is an answer
set for program~(\ref{ex})
%\beq
%\ba l
%p \lar \no\ \no\ p\\
%p \lar p,q
%\ea
%\eeq{ex}
since it is the answer set for the reduct (\ref{ex-r})
of (\ref{ex}) relative to $\{p\}$.

We say that a formula (or a program) is {\sl normal} if it does not
contain classical negation.  We will sometimes identify a normal formula $F$
with the propositional formula obtained from $F$ by replacing every comma with
$\wedge$, every semicolon with $\vee$, and every occurrence of \no\ with $\neg$.
It is easy to see that, for any normal formula~$F$ and any set $X$ of atoms,
$X\satisfies F$ iff $X$ satisfies~$F$ in the sense of propositional logic.

\section{Supported Sets and Completion} \label{sec:supportedness}

This section is a review of supported sets and completion for the class of
logic programs introduced above.

We say that a set $X$ of literals is {\sl supported by} a program $\Pi$
if for every literal $L \in X$ there exists a rule~(\ref{rule}) in $\Pi$
such that $\ii{Head}=L$ and \hbox{$X \satisfies \ii{Body}$}.\footnote{In
\cite{bar94} and~\cite{ino98} the definition of supportedness from~\cite{apt88}
is generalized in a different direction---it is extended to disjunctive
programs.}
For instance, each of the sets $\{p\}$, $\{\neg q\}$ is supported by the
program
\beq
\ba l
p \lar \no\ \neg q \\
\neg q \lar \no\ p
\ea
\eeq{ex:p-or-q}
but their union  is not.  The set $\{p\}$ is supported by program~(\ref{ex});
$\{q\}$ is not.

We are interested in the relationship between the concept of an answer set on
the one hand, and the conjunction of the closure and supportedness conditions
on the other.  According to the
following proposition, every answer set is both closed under the program
and supported by it.

\begin{prop} \label{easy-main}
For any program $\Pi$
and any consistent set $X$ of literals, if $X$ is an answer set
for~$\Pi$ then $X$ is closed under and supported by~$\Pi$.
\end{prop}

For instance, the answer sets for program~(\ref{ex:p-or-q}) are $\{p\}$ and
$\{\neg q\}$; each of them is closed under and supported by~(\ref{ex:p-or-q}).
The answer sets for program~(\ref{ex}) are $\emptyset$ and $\{p\}$; each of
them is closed under and supported by~(\ref{ex}).

The converse, in general, is not true.  The easiest counterexample is the
program consisting of one rule $p \lar p$.
The set $\{p\}$ is closed under this program and supported by it,
although it is not an answer set.
Tightness, defined in Section~\ref{sec:tightness} below, is a syntactic
condition that eliminates programs like this.

Let $\Pi$ be a finite normal program.  The ``completion'' of~$\Pi$ is the set
of propositional formulas defined as follows. If $A$ is an atom or the
symbol $\bot$, by $\ii{Comp}(\Pi,A)$ we denote the propositional formula
\beq
A \equiv \ii{Body}_1 \vee \cdots \vee \ii{Body}_k
\eeq{comp}
where the disjunction extends over all rules
\beq
A \lar \ii{Body}_i
\eeq{comp-rule}
in $\Pi$ with the head $A$. The {\sl
completion} of $\Pi$ is the set of formulas $\ii{Comp}(\Pi,A)$ for all
$A$.\footnote{This is essentially the definition from \cite{llo84a}
restricted to the propositional case.  It is restricted to
finite programs to avoid the need to use an infinite disjunction
in~(\ref{comp}).}

 For instance, the bodies of rules (\ref{ex}), written as
propositional formulas, are $\neg\neg p$ and $p \wedge q$; for this program
$\Pi$, formulas~(\ref{ex-comp})
%\beq
%\ba l
%p \equiv \neg \neg p \vee (p \wedge q), \\
%q \equiv \bot
%\ea
%\eeq{ex-comp}
are $\ii{Comp}(\Pi,p)$ and
$\ii{Comp}(\Pi,q)$. In addition to these two formulas, the completion of
this program includes also $\ii{Comp}(\Pi,\bot)$, which is the tautology
$\bot \equiv \bot$.

In application to
finite normal programs, the conjunction of closure and supportedness
exactly corresponds to the program's completion:

\begin{prop} \label{cs}
For any finite normal program $\Pi$, a set of atoms
satisfies the completion of $\Pi$ iff it is closed under and supported
by~$\Pi$.
\end{prop}

 From Propositions~\ref{easy-main} and~\ref{cs}, we conclude:

\begin{cor} \label{cor:comp}
For any finite normal program $\Pi$ and any set $X$ of atoms,
if $X$ is an answer set for $\Pi$ then $X$ satisfies the completion of $\Pi$.
\end{cor}

\section{Tight Programs} \label{sec:tightness}

To define the concept of a tight program, we need a few auxiliary
definitions.

Recall that an occurrence of a formula~$F$ in a formula~$G$ is {\sl singular}
if the symbol before this occurrence is $\neg$; otherwise, the occurrence is
{\sl regular}~\cite{lif99d}.
It is clear that the occurrence of $F$ can be singular
only if $F$ is an atom. For any formula $G$, by $\ii{lit}(G)$ we denote the set
of all literals having regular occurrences in $G$. For instance,
$\ii{lit}(p; \no\ \neg r) = \{p,\neg r\}$.
For any formula $G$, by $\ii{poslit}(G)$ we denote the set of all literals
having a regular occurrence in $G$ that is not in the scope of negation as
failure.  For instance,
$\ii{poslit}(p, \no\ q, (\no\ \no\ r; \neg s)) = \{p,\neg s\}$.

For any program $\Pi$ and any set $X$ of literals, we say
about literals $L,L'\in X$ that $L$ is a {\sl parent} of $L'$ relative to
$\Pi$ and $X$ if there is a rule (\ref{rule}) in $\Pi$ such that
\begin{itemize}
\item
$X \satisfies \ii{Body}$,
\item
$L \in \ii{poslit}(\ii{Body})$, and
\item
$L'=\ii{Head}$.
\end{itemize}
For instance, the parents of $p$ relative to the program
\beq
\ba l
p \lar \no\ q\\
q \lar \no\ p\\
p \lar p, r
\ea
\eeq{ex2}
and the set $\{p,q,r\}$
are $p$ and $r$; on the other hand, $p$ has no parents relative to
(\ref{ex2}) and the set $\{p,q\}$.

Now we are ready to give the main definition of this paper:

A program $\Pi$  is {\sl tight} on a set $X$ of
literals if there is no
infinite sequence $L_1,L_2,\dots$ of elements of $X$ such that for every
$i$, $L_{i+1}$  is a parent of $L_i$ relative to $\Pi$ and $X$.

In other words, $\Pi$  is tight on a set $X$
iff the parent relation
relative to $\Pi$ and $X$ is well-founded.\footnote{A binary relation $R$ is
{\sl well-founded} if there is no infinite sequence $x_1,x_2,\dots$ of elements
of its domain such that, for all $i$, $x_{i+1}Rx_i$.}

If $X$ is finite then the tightness condition can be reformulated as
follows: there is no finite sequence $L_1,\dots,L_n$ of elements of $X$
($n>1$) such that for every $i$ ($1\leq i<n$), $L_{i+1}$ is a parent of $L_i$
relative to $\Pi$ and $X$, and $L_n=L_1$.

For instance, program~(\ref{ex})
is tight on  $\{p\}$: $p$ does not have parents relative to~(\ref{ex})
and~$\{p\}$, so that
the parent relation relative to~(\ref{ex}) and $\{p\}$ is well-founded.  But
that program is not tight on $\{p,q\}$.  Indeed, $p$ and $q$ are the parents
of $p$ relative to~(\ref{ex}) and $\{p,q\}$, so that in the sequence
$p,p,\dots$ every element is followed by its parent.

The proposition below gives an equivalent characterization of tightness:

\begin{prop} \label{prop2}
A program $\Pi$ is tight on
a set $X$ of literals iff there exists a function $\lambda$ from $X$ to
ordinals satisfying the following condition:
\begin{itemize}
\item[(*)] for every rule~(\ref{rule}) in $\Pi$ such that $\ii{Head}\in X$
and $X \satisfies \ii{Body}$, and  for every $L$ in $X \cap
\ii{poslit}(\ii{Body})$, $\lambda(L) < \lambda(\ii{Head}).$
\end{itemize}
\end{prop}

For instance, to show that program~(\ref{ex}) is tight on $\{p\}$, we can
take $\lambda(p) = 0$.  To show that program
\beq
\ba l
p \lar q;\no\ r
\ea
\eeq{ex1}
is tight on $\{p,q,r\}$, take $\lambda(p) = 1$, $\lambda(q) = \lambda(r) = 0$.

If $X$ is finite then the values of $\lambda$ in the statement of
Proposition~\ref{prop2} can be assumed to be finite.

Proposition~\ref{prop2} is a special case of the following general fact:

\begin{proof*}[Lemma from Set Theory]
A binary relation $R$ is well-founded iff there exists a function $\lambda$
from the domain of~$R$ to ordinals such that, for all $x$ and $y$, $xRy$
implies $\lambda(x)<\lambda(y)$.
\end{proof*}

To compare the definition of tightness above with the definition given earlier
in~\cite{bab00}, assume that the rules of~$\Pi$ have the form
\beq
\ii{Head} \lar L_1,\dots,L_m,\no\ L_{m+1},\dots,\no\ L_n
\eeq{rule2}
where each $L_i$ is a  literal.  In this case, condition~(*) says:
{\it for every rule~(\ref{rule2}) in $\Pi$, if
\beq
\ii{Head} \in X,
\eeq{b1}
\beq
L_1,\dots,L_m \in X
\eeq{b2}
and
\beq
L_{m+1},\dots,L_n \not\in X
\eeq{b3}
then, for all $L \in X \cap \{L_1,\dots,L_m\}$,
$\lambda(L) < \lambda(\ii{Head})$.}
In view of~(\ref{b2}), the intersection $X \cap \{L_1,\dots,L_m\}$ here can be
replaced by $\{L_1,\dots,L_m\}$.  The only difference between this form of
condition~(*) and the corresponding condition in~\cite{bab00} is the
presence of restriction~(\ref{b3}).
%\footnote{\label{ftn}The possibility
% of including this restriction was suggested to us by Hudson Turner on May 2,
% 2001.}
The additional generality gained by including~(\ref{b3}) can be
illustrated by the program
$$
\ba l
p\\
q\\
p \lar p, \no\ q
\ea
$$
---it is tight on $\{p,q\}$ in the sense of this paper, but not
in the sense of \cite{bab00}.

\section{Absolutely Tight Programs}\label{sec:abs}

The following modification of the tightness condition is often useful.
A program $\Pi$  is {\sl absolutely tight} if there is no infinite sequence
$L_1,L_2,\dots$ of literals such that for every $i$ there is a rule
(\ref{rule}) in $\Pi$ for which $L_{i+1} \in \ii{poslit}(\ii{Body})$ and
$L_i=\ii{Head}$.  It is clear that an absolutely tight program is
tight on any set of literals.

To prove that a program~$\Pi$ is absolutely tight, it is sufficient to
find a function~$\lambda$ from literals to ordinals such that for every
rule~(\ref{rule}) in $\Pi$ with $\ii{Head}\neq\bot$ and for every literal
$L \in \ii{poslit}(\ii{Body})$, $\lambda(L) < \lambda(\ii{Body})$.

For a program containing finitely many atoms, absolute tightness can be
characterized as follows.  The {\sl positive dependency graph} of a
program $\Pi$ is the directed graph $G$ such that
\begin{itemize}
\item the vertices of $G$ are the literals that have regular occurrences
in $\Pi$, and
\item $G$ has an edge from $L$ to $L'$ if there is a rule (\ref{rule})
in $\Pi$ for which $L \in \ii{poslit}(\ii{Body})$ and $L'=\ii{Head}$.
\end{itemize}
A program containing finitely many atoms is absolutely tight iff
its positive dependency graph has no cycles.

In application to programs whose rules have the form~(\ref{rule2}) and
contain neither~$\bot$ nor classical negation,
the definition of absolute tightness
above turns into (the propositional case of) Fages' original definition of
tightness~\cite{fag94}.

\section{Generalization of Fages' Theorem}\label{sec:thm}

\begin{thm} \label{fages-extended}
For any program $\Pi$ and any
consistent set $X$ of literals such that $\Pi$ is tight on $X$,
$X$ is an answer set for~$\Pi$ iff $X$ is closed under and supported by~$\Pi$.
\end{thm}

For instance, program~(\ref{ex:p-or-q}) is tight on the sets~$\{p\}$
and~$\{\neg q\}$ that are closed under and supported by~(\ref{ex:p-or-q}).
By Proposition~\ref{easy-main} and the theorem above, it follows that
$\{p\}$ and $\{\neg q\}$ are the answer sets for~(\ref{ex:p-or-q}).

By Proposition~\ref{cs}, we conclude:

\begin{cor} \label{main-cor}
For any finite normal program $\Pi$ and any set $X$ of atoms such that $\Pi$
is tight on~$X$,
$X$ is an answer set for~$\Pi$ iff $X$~satisfies the completion of~$\Pi$.
\end{cor}

For instance, program~(\ref{ex}) is tight on the models $\emptyset$, $\{p\}$
of its completion~(\ref{ex-comp}).  In accordance with
Proposition~\ref{easy-main} and
Corollary~\ref{main-cor}, these two models are the answer sets
for~(\ref{ex}).

By $\ii{pos}(\Pi)$ we denote the set of literals~$L$ such that
$\Pi$ contains a rule~(\ref{rule}) with $\ii{Head} \neq \bot$ and
$L\in \ii{poslit}(\ii{Body})$.  For instance, if $\Pi$ is (\ref{ex}) then
$\ii{pos}(\Pi) = \{p,q\}$.  If a set~$X$ of literals is disjoint from
$\ii{pos}(\Pi)$ then no literal in $X$ has a parent relative to $\Pi$ and $X$,
and consequently $\Pi$ is tight on $X$.  We conclude:

\begin{cor} \label{fages-pos}
For any program $\Pi$ and any consistent set $X$ of literals disjoint from
$\ii{pos}(\Pi)$, $X$ is an answer set for~$\Pi$ iff $X$ is closed under and
supported by~$\Pi$.
\end{cor}

By Proposition~\ref{cs}, it follows then:

\begin{cor} \label{cor-fages-pos}
For any finite normal program $\Pi$ and any set $X$ of atoms disjoint from
$\ii{pos}(\Pi)$, $X$ is an answer set for~$\Pi$ iff $X$ satisfies the
completion of~$\Pi$.
\end{cor}

Since an absolutely tight program is tight on every set of literals, we
conclude from Theorem~\ref{fages-extended} and Corollary~\ref{main-cor}:

\begin{cor} \label{fages-extended-abs}
For any absolutely tight program $\Pi$ and any consistent set $X$ of literals,
$X$ is an answer set for~$\Pi$ iff $X$ is closed under and supported by~$\Pi$.
\end{cor}

\begin{cor} \label{main-cor-abs}
For any finite normal absolutely tight program $\Pi$ and any set $X$ of atoms,
$X$ is an answer set for~$\Pi$ iff $X$~satisfies the completion of~$\Pi$.
\end{cor}

\section{Example: The N-Queens Problem} \label{sec:eightqueens}

In the $n$-queens problem, the goal is to find a configuration of $n$ queens
on an $n \times n$ chessboard such that no queen can be taken by any
other queen. In other words, (a) no two queens may be on the same column,
(b) no two queens may be on the same row, and (c) no two queens may be
on the same diagonal.

A solution to the $n$-queens problem can be described by a set of
atoms of the form $\ii{queen}(R,C)$ ($1\leq R,C\leq n$) satisfying
conditions (a)--(c); including
$\ii{queen}(R,C)$ in the set indicates that there is a queen in
position~$(R,C)$.

The $n$-queens problem can be described by a program
whose answer sets are solutions, as follows.  The selections satisfying
condition~(a) correspond to the answer sets for the program consisting of
the rules
\beq
\ii{queen}(R,C) \lar  \no\ \no\ \ii{queen}(R,C)
\eeq{nested-queen1}
for all $R,C$ in $\{1,\dots,n\}$,
\beq
\lar  \no\ \ii{queen}(1,C), \dots, \no\ \ii{queen}(n,C)
\eeq{nested-queen2}
for all $C$ in $\{1,\dots,n\}$, and
\beq
\lar  \ii{queen}(R,C), \ii{queen}(R1,C)
\eeq{nested-queen3}
for all $R,R1,C$ in $\{1,\dots,n\}$ such that $R < R1$.
Conditions~(b) and~(c) are represented by the constraints
\beq
\lar \ii{queen}(R,C), \ii{queen}(R,C1)
\eeq{column-constraint}
for all $R,C,C1$ in $\{1,\dots,n\}$ such that $C < C1$ and
\beq
\lar \ii{queen}(R,C), \ii{queen}(R1,C1)
\eeq{diagonal-constraint}
for all $R,R1,C,C1$ in $\{1,\dots,n\}$ such that $C < C1$
and \hbox{$|R - R1| = |C - C1|$.}

The answer sets for program (\ref{nested-queen1})--(\ref{diagonal-constraint})
are in a 1--1 correspondence with the possible arrangements of
$n$ queens.  According to~\cite{fer03},
rules~(\ref{nested-queen1})--(\ref{nested-queen3}) can be
rewritten as weight constraints
\beq
1\{\ii{queen}(1,C),\dots,\ii{queen}(n,C)\}1
\eeq{choice-queen}
for all $C$ in $\{1,\dots,n\}$
Then the program can be
presented to {\sc smodels} as shown in Figure~\ref{fig:smodels}.

\begin{figure}
\begin{verbatim}
number(1..n).

1{queen(R,C) : number(R)}1 :- number(C).

:- queen(R,C), queen(R,C1), number(R;C;C1), C < C1.

:- queen(R,C), queen(R1,C1), number(R;C;R1;C1),
   C < C1, abs(R-R1) == abs(C-C1).
\end{verbatim}
\caption{An {\sc smodels} representation of the $n$-queens problem.}
\label{fig:smodels}
\end{figure}

Program (\ref{nested-queen1})--(\ref{diagonal-constraint})  is
a finite normal absolutely tight program.
By Corollary~\ref{main-cor-abs}, its answer sets are identical to the models
of its completion.
(Note the use of nested negations in rule~(\ref{nested-queen1}); this is
the reason why the new generalization of Fages' theorem is needed here.)
This fact can be used to find solutions to the n-queens problem using
{\sc cmodels}, that is to say, by running
a satisfiability solver on the program's completion. {\sc cmodels}
transforms the input shown in Figure~\ref{fig:smodels} into
(\ref{nested-queen1})--(\ref{diagonal-constraint}), computes the
completion, clausifies it, and calls {\sc mchaff} to find a model.
For $n=20$ {\sc cmodels} finds a
solution in 2 seconds (for comparison, {\sc smodels}, given the same input
file, finds one in 55 seconds).\footnote{
We have used {\sc lparse 1.0.11}, {\sc smodels 2.27}, {\sc cmodels 1.03},
and {\sc mchaff spelt3}.
All CPU times here are for a
SunBlade 1000, with two 600MHz UltraSPARC-III processors and 5GB RAM.}
For $n=25$,  {\sc cmodels} finds a
solution in 3 seconds, whereas {\sc smodels} requires more than 2 hours.

This example confirms the conjecture underlying the design of {\sc cmodels}:
using satisfiability solvers to compute answer sets
for tight programs may be computationally advantageous.  Systematic
experimental evaluation of this form of answer set programming is a topic
for future research.

\section{Transitive Closure} \label{sec:tc}

In logic programming, the transitive closure $\ii{tc}$ of a binary
predicate $p$ is usually defined by the rules
%\beq
$$
\ba l
\ii{tc}(x,y) \lar p(x,y)\\
\ii{tc}(x,y) \lar p(x,v), \ii{tc}(v,y).
\ea
$$
%\eeq{trans-p}
If we combine this definition \ii{Def} with any set $\Pi$ of facts
defining $p$, and consider the minimal model of the resulting
program, the extent of $\ii{tc}$ in this model will be the transitive closure of
the extent of $p$.  In this sense, \ii{Def} is a correct characterization
of the concept of transitive closure.  We know,
on the other hand, that the sets of atoms closed under and supported by
$\Pi\cup\ii{Def}$ may be different from the
minimal model.  In these ``spurious'' sets of atoms, $\ii{tc}$ is weaker
than the transitive closure of $p$. The absence of such ``spurious'' sets
can be assured by requiring that facts in~$\Pi$ define relation~$p$ to be
acyclic.

In this section we study the more general situation when $\Pi$ is a logic
program, not necessarily a set of facts.  This program may define
several predicates besides $p$.  Even $\ii{tc}$ is allowed to occur in $\Pi$,
except that all occurrences of this predicate are supposed to be in the
bodies of rules, so that all rules defining $\ii{tc}$ in $\Pi\cup\ii{Def}$ will
belong to \ii{Def}.  The rules of $\Pi$ may include negation as failure,
and, accordingly, we talk about answer sets
instead of the minimal model.  Program $\Pi\cup\ii{Def}$ may have
many answer sets.  According to Proposition~\ref{thm1} below,
the extent of~$\ii{tc}$ in each of these sets is the transitive closure of
the extent of~$p$ in the same set.

Recall that programs in the sense of Section~\ref{sec:programs} are
propositional objects; there are no variables in them.  Expressions
containing variables, such as~\ii{Def}, can be treated as
schematic:  we select a non-empty set $C$ of symbols (``object constants'')
and view an expression with variables as shorthand for the set of all its
ground instances obtained by substituting these symbols for variables.
It is convenient, however, to be a little more general.  We assume
$p$ and $\ii{tc}$ to be functions from $C\times C$ to the set of
atoms such that all atoms $p(x,y)$ and $\ii{tc}(x,y)$ are pairwise distinct.

\begin{prop}\label{thm1}
Let $\Pi$ be a program that does not contain atoms of the form $\ii{tc}(x,y)$ in
the heads of rules.  If $X$ is an answer set for
$\Pi \cup \ii{Def}$
then
\beq
\{\langle x,y \rangle: \ii{tc}(x,y) \in X\}
\eeq{tc}
is the transitive closure of
\beq
\{\langle x,y \rangle: p(x,y) \in X\}.
\eeq{r}
\end{prop}

If atoms of the form $\ii{tc}(x,y)$ do not occur in $\Pi$ at all then the
answer sets for $\Pi\cup\ii{Def}$ are actually in a 1-1 correspondence
with the answer sets for $\Pi$.\footnote{This observation is due to Hudson
Turner (personal communication, October 3, 2000).} The answer set for
$\Pi\cup\ii{Def}$ corresponding to an answer set $X$ for $\Pi$ is obtained
from $X$ by adding a set of atoms of the form $\ii{tc}(x,y)$.
% This is easy
% to prove using the splitting set theorem \cite{lif94e}.

Under what conditions can we assert that the consistent sets of literals
closed under and supported by a program containing \ii{Def} are not
``spurious''?  As we know from Theorem~\ref{fages-extended}, such a condition
is provided by the tightness of the program.  The verification of the
tightness of programs containing \ii{Def} is facilitated by the theorem below,
which tells us that in some cases the tightness of a program is preserved after
adding~\ii{Def} to it.

For any program $\Pi$ and any set $X$ of literals, we say
about literals $L,L'\in X$ that $L'$ is an {\sl ancestor} of $L$ relative to
$\Pi$ and $X$ if there exists a finite sequence of literals
$L_1,\dots,L_n\in X$ $(n>1)$ such that $L=L_1$, $L'=L_n$ and for every~$i$
($1\leq i<n$), $L_{i+1}$ is a parent of $L_i$ relative to $\Pi$ and $X$.
In other words, the ancestor relation is the transitive closure of the
parent relation.

\begin{thm}\label{thm2}
Let $\Pi$ be a program that does not contain atoms of the form $\ii{tc}(x,y)$ in
the heads of rules.  For any set $X$ of literals, if
\begin{enumerate}
\item[(i)] $\Pi$ is tight on $X$,
\item[(ii)] $\{\langle x,y \rangle:\ p(y,x) \in X\}$ is well-founded, and
\item[(iii)] no atom of the form $\ii{tc}(x,y)$ is an ancestor of an atom of the
form $p(x,y)$ relative to $\Pi$ and $X$,
\end{enumerate}
then
$\Pi \cup \ii{Def}$ is tight on $X$.
\end{thm}

By Theorem~\ref{fages-extended} and Proposition~\ref{thm1}, we conclude:

\begin{cor} \label{cor:thm2a}
Let $\Pi$ be a program that does not contain atoms of the form $\ii{tc}(x,y)$
in the heads of rules, and let~$X$ be a set of literals satisfying
conditions (i)--(iii) from Theorem~\ref{thm2}.  If, in addition,
\begin{enumerate}
\item[(iv)] $X$ is a consistent set closed under and supported by
$\Pi \cup \ii{Def}$
\end{enumerate}
then $X$ is an answer set for $\Pi \cup \ii{Def}$, and
$$\{\langle x,y \rangle: \ii{tc}(x,y) \in X\}$$
is the transitive closure of
$$\{\langle x,y \rangle: p(x,y) \in X\}.$$
\end{cor}

Proposition~\ref{cs} shows that if $X$ is a set of atoms and $\Pi$ is a
finite program without classical negation then condition~(iv) can be
reformulated as follows:
\begin{enumerate}
\item[(iv$'$)]
$X$ is a model of the completion of~$\Pi \cup \ii{Def}$.
\end{enumerate}

Condition~(ii) in the statement of Theorem~\ref{thm2} is similar to the
acyclicity property mentioned at the beginning of this section.
In fact, if the underlying set $C$ of constants is finite then
(ii) is obviously equivalent to the following condition: there is no finite
sequence $x_1,\dots,x_n\in C$ $(n>1)$ such that
\beq
p(x_1,x_2),\dots,p(x_{n-1},x_n)\in X
\eeq{finite-seq}
and $x_n=x_1$.  For an infinite $C$, well-foundedness implies acyclicity,
but not the other way around.

Here is a useful syntactic sufficient condition for~(ii):

\begin{prop}\label{prop1}
If $\Pi$ contains constraint
\beq
\lar \ii{tc}(x,x)
\eeq{irref-tc}
and $C$ is finite then, for every set $X$ of literals closed under
$\Pi\cup\ii{Def}$, set $\{\langle x,y \rangle:\ p(y,x) \in X\}$ is
well-founded.
\end{prop}

Without condition~(ii), the assertion of the theorem would be incorrect.
Program $\Pi$ that consists of one fact $p(1,1)$, with $C = \{1,2\}$ and
$$X=\{p(1,1), \ii{tc}(1,1), \ii{tc}(1,2)\},$$
provides a counterexample.

Condition~(iii) is essential as well.  Indeed, take $\Pi$ to be
\bdm
\ba l
p(x,y) \lar \ii{tc}(x,y).
\ea
\edm
With $C = \{1,2\}$, set $X=\{p(2,1),\ii{tc}(2,1)\}$ is closed under and supported
by~$\Pi\cup\ii{Def}$, but is is not an answer set for $\Pi\cup\ii{Def}$:
the only answer set for this program is empty.

\section{Example: The Blocks World}\label{sec:bw}

As an example of the use of Theorem~\ref{thm2}, consider a
``history program'' for the blocks world---a program whose answer sets
represent possible ``histories''of the blocks world over a fixed time
interval.   A history of
the blocks world is characterized by the truth values of atoms of two
kinds: $\ii{on}(b,l,t)$ (``block $b$ is on location $l$ at time~$t$'') and
$\ii{move}(b,l,t)$ (``block $b$ is moved to location $l$ between times $t$
and $t+1$''). Here
\begin{itemize}
\item
$b$ ranges over a finite set of ``block constants,''
\item
$l$ ranges over the set of location constants that consists of the block
constants and the constant \ii{table},
\item
$t$ ranges over the symbols representing an initial segment of integers
$0,\dots,T$,
\end{itemize}
except that in $\ii{move}(b,l,t)$ we require $t<T$.  One other kind of
atoms used in the program is $\ii{above}(b,l,t)$:
``block $b$ is above location $l$ at time~$t$''.  These atoms are used
to express constraint~(\ref{hp5a}) that requires every block to be
``supported by the table'' and thus eliminates stacks of blocks flying
in space.

The program consists of the following rules:\footnote{\label{ftnt}This
program is similar
to the history program for the blocks world from~\cite{lif99c}.  Instead of
rules~(\ref{hp1}), the program in that paper contains a pair of
disjunctive rules; according to Theorem~1 from~\cite{erd99}, this difference
does not affect the program's answer sets.  The intuitive meaning of rules
(\ref{hp2})--(\ref{hp3h}) is discussed in~\cite{lif99c}, Section~6.
Rule~(\ref{hp3a}) prohibits concurrent actions.  Rules (\ref{hp4}) and
(\ref{hp5}) were suggested to us by Norman McCain and Hudson Turner on
June 11, 1999; similar rules are discussed in \cite{lif99c}, Section~8.}
\medskip
\beq
\ba l
\ii{on}(b,l,0) \lar \no\ \neg \ii{on}(b,l,0)\\
\neg \ii{on}(b,l,0) \lar \no\ \ii{on}(b,l,0)\\
\ii{move}(b,l,t) \lar \no\ \neg \ii{move}(b,l,t) \\
\neg \ii{move}(b,l,t) \lar \no\ \ii{move}(b,l,t) \\
\ea
\eeq{hp1}
\beq
\ba l
\ii{on}(b,l,t+1) \lar \ii{move}(b,l,t) \\
\ii{on}(b,l,t+1) \lar \ii{on}(b,l,t), \no\ \neg \ii{on}(b,l,t+1) \\
\ea
\eeq{hp2}
\beq
\neg \ii{on}(b,l,t) \lar \ii{on}(b,l',t) \qquad (l \neq l') \\
\eeq{hp2a}
\beq
\lar \ii{on}(b,b'',t), \ii{on}(b',b'',t) \qquad (b \neq b')
\eeq{hp3}
\beq
\lar \ii{move}(b,l,t), \ii{on}(b',b,t) \\
\eeq{hp3h}
\beq
\lar \ii{move}(b,l,t), \ii{move}(b',l',t) \qquad
                              (b \neq b'\hbox{ or } l\neq l') \\
\eeq{hp3a}
\beq
\ba l
\ii{above}(b,l,t) \lar \ii{on}(b,l,t) \\
\ii{above}(b,l,t) \lar \ii{on}(b,b',t), \ii{above}(b',l,t) \\
\ea
\eeq{hp4}
\beq
\lar \ii{above}(b,b,t)
\eeq{hp5}
\beq
\lar \no\ \ii{above}(b,table,t)
\eeq{hp5a}

To illustrate the use of Theorem~\ref{thm2}, in Section~\ref{sec:proofs} we
use it to prove the following proposition:

\begin{prop}\label{prop3}
Program~(\ref{hp1})--(\ref{hp5a}) is tight on every set of literals that is
closed under it.
\end{prop}

This proposition, in combination with Proposition~\ref{easy-main} and
Theorem~\ref{fages-extended}, tells us
that the answer sets for~(\ref{hp1})--(\ref{hp5a}) can be characterized as
the sets that are closed under this program and supported by it.  These
answer sets can be computed by eliminating classical negation in favor of
new atoms and generating models of the completion of the resulting program.

The idea of the proof is to check first that program~(\ref{hp1})--(\ref{hp3a}),
(\ref{hp5}), (\ref{hp5a}) is tight, and then use Theorem~\ref{thm2} to
conclude that tightness is preserved when we add the definition~(\ref{hp4}) of
\ii{above}.  There are two complications, however, that need to be taken into
account.

First, \ii{on} and \ii{above} are ternary predicates, not binary.  To relate
them to the concept of transitive closure, we can
say that any binary ``slice'' of \ii{above} obtained by fixing
its last argument is the transitive closure of the corresponding ``slice'' of
\ii{on}.  Accordingly, Theorem~\ref{thm2} will need to be applied $T+1$ times,
once for each slice.

Second, the first two arguments of \ii{on} do not come from the same set $C$
of object constants, as required in the framework of Theorem~\ref{thm2}: the
set of block constants is a proper part of the set of location constants.  In
the proof, we will introduce a program similar to (\ref{hp1})--(\ref{hp5a})
in which,
syntactically, \ii{table} is allowed as the first argument of both \ii{on} and
\ii{above}.

\section{Proofs} \label{sec:proofs}

\begin{lem} \label{monoton}
Given a formula $F$ without negation as failure and two sets $Z,Z'$
of literals such that $Z' \subseteq Z$, if $Z' \satisfies F$ then $Z
\satisfies F$.
\end{lem}

\begin{proof*} Immediate by structural induction.
\end{proof*}

\medskip
The following lemma is the special case of Proposition~\ref{easy-main} in
which $\Pi$ is assumed to be a program without negation as failure.

\begin{lem} \label{left-to-right}
For any program $\Pi$ without negation as failure and any
consistent set $X$ of literals, if $X$~is an answer set for~$\Pi$ then
$X$~is closed under and supported by~$\Pi$.
\end{lem}

\begin{proof*}
Let $\Pi$ be a program without negation as failure and $X$
be an answer set for $\Pi$. By the definition of an answer set for
programs without negation as failure, $X$ is closed under $\Pi$. To prove
supportedness, take any literal~$L$ in~$X$.  Since $X$ is minimal among
the sets closed under $\Pi$, $X \setminus \{L\}$ is not closed under
$\Pi$. This means that $\Pi$ contains a rule (\ref{rule}) such that $X
\setminus \{L\} \satisfies \ii{Body}$ but $\ii{Head} \not\in X \setminus
\{L\}$. By Lemma~\ref{monoton}, $X \satisfies \ii{Body}$. Since $X$ is
closed under $\Pi$, it follows that $\ii{Head} \in X$, so that $\ii{Head}
= L$.
\end{proof*}

The definition of the reduct $F^X$ of a formula $F$ is similar to the
definition of the reduct of a program given in Section~\ref{sec:programs}.

\begin{lem} \label{closure-supportedness}
For any formula $F$, any program $\Pi$, and any consistent
set $X$ of literals,
\begin{itemize}
\item[(i)] $X \satisfies F$ iff $X \satisfies F^X$;
\item[(ii)] $X$ is closed under $\Pi$ iff $X$ is closed under $\Pi^X$;
\item[(iii)] $X$ is supported by $\Pi$ iff $X$ is supported by $\Pi^X$.
\end{itemize}
\end{lem}

\begin{proof*}
Part (i) is immediate by structural induction; parts
(ii) and (iii) follow from (i).
\end{proof*}

\begin{proof*}[Proof of Proposition~\ref{easy-main}
(Section~\ref{sec:supportedness})]
Consider a program $\Pi$ and an answer set $X$ for~$\Pi$.
By the definition of an answer set, $X$ is an answer set for $\Pi^X$.
Then, by Lemma~\ref{left-to-right}, $X$ is closed under and supported by
$\Pi^X$. By Lemma~\ref{closure-supportedness}(ii,iii), it follows that
$X$ is closed under and supported by $\Pi$.
\end{proof*}

\begin{proof*}[Proof of Proposition~\ref{cs}
(Section~\ref{sec:supportedness})]
Let $\Pi$ be a finite normal program. Recall that the
completion of $\Pi$ consists of the equivalences (\ref{comp}) where $A$ is an
atom or the symbol $\bot$. It is clear that a set $X$ of atoms satisfies the
completion of $\Pi$ iff, for each~$A$,
\begin{itemize}
\item[(a)] for every rule (\ref{comp-rule}) in $\Pi$ with the head $A$, if $X
\satisfies \ii{Body}_i$ then $A \in X$, and
\item[(b)] if $A \in X$ then $X \satisfies \ii{Body}_i$
for some rule (\ref{comp-rule}) in $\Pi$ with the head $A$.
\end{itemize}
Condition (a) expresses that
$X$ is closed under $\Pi$, and condition (b) expresses that $X$ is
supported by $\Pi$.
\end{proof*}

\begin{proof*}[Proof of Lemma from Set Theory
(Section~\ref{sec:tightness})]
The ``if'' part follows from the
well-foundedness of $<$ on sets of ordinals.  To prove the ``only if'' part,
consider the following transfinite sequence of subsets of the domain of $R$:
$$
\ba l
S_0 = \emptyset, \\
S_{\alpha+1} = \{x:\ \forall y (yRx \Rightarrow y \in S_{\alpha})\}, \\
S_{\alpha} = \bigcup_{\beta < \alpha} S_{\beta}\quad
                \hbox{if $\alpha$ is a limit ordinal.}
\ea
$$
For any $x\in\bigcup_\alpha S_\alpha$, define $\lambda(x)$ to be the
smallest $\alpha$ such that $x\in S_\alpha$.  From the well-foundedness
of $R$ we can conclude that
$\bigcup_\alpha S_\alpha$ is the whole domain of $R$.
\end{proof*}

\begin{lem} \label{aux}
For any formula $F$ and any set $X$ of literals,
$X \satisfies F$ iff $X \cap \ii{lit}(F) \satisfies F$.
\end{lem}

\begin{proof*}
Immediate by structural induction.
\end{proof*}

\medskip
The following lemma is the special case of one half of
Theorem~\ref{fages-extended} in which
$\Pi$ is assumed to be a program without negation as failure.

\begin{lem} \label{right-to-left}
Let $\Pi$ be a program without negation as failure. For any consistent set $X$ of literals such that $\Pi$
is tight on $X$, if $X$~is closed under  and supported by~$\Pi$ then
$X$~is an answer set for~$\Pi$.
\end{lem}

\begin{proof*}
Let $\Pi$ be a program without negation as failure
and $X$ be a consistent set of literals such that $\Pi$
is tight on $X$.  Suppose that $X$~is closed under  and supported by~$\Pi$.
By the definition of an answer set for programs without negation as
failure, we need to show that no proper subset of $X$ is closed under
$\Pi$. Let $Y$ be a proper subset of $X$.  Note first that there exists a
literal $L$ in $X\setminus Y$ that does not have a parent in $X \setminus Y$
relative to $\Pi$  and $X$. Indeed, assume that there is no such literal, so
that every literal in $X\setminus Y$ has a parent relative to $\Pi$ and $X$
in this set; then there exists an infinite sequence $L_1,L_2,\dots$ of
elements of $X\setminus Y$ such that $L_{i+1}$ is a parent of $L_i$, which
contradicts the assumption that $\Pi$ is tight on $X$.

Take such a literal~$L$.  Since $X$ is supported by $\Pi$, there is a rule
$$ L \lar \ii{Body}$$
in $\Pi$ such that
\beq
X \satisfies \ii{Body}.
\eeq{xpn}
By the definition of the parent relation, the elements of $X \cap \ii{poslit}(\ii{Body})$
are parents of $L$ relative to $\Pi$ and $X$.
By the choice of~$L$, no parent of $L$ relative to $\Pi$ and $X$
belongs to $X \setminus Y$, so that $X \cap \ii{poslit}(\ii{Body})$ is disjoint
from $X \setminus Y$. Consequently,
$X \cap \ii{poslit}(\ii{Body}) \subseteq Y$. Since
$\Pi$ does not contain negation as failure,
$\ii{lit}(\ii{Body}) = \ii{poslit}(\ii{Body})$, so
that
\beq
X \cap \ii{lit}(\ii{Body}) \subseteq Y.
\eeq{xpny}
By Lemma~\ref{aux}, we can conclude from (\ref{xpn}) that
$$X \cap \ii{lit}(\ii{Body}) \satisfies \ii{Body}.$$
In view of (\ref{xpny}), it follows by Lemma~\ref{monoton} that $Y
\satisfies \ii{Body}$.   Since~$L\notin Y$, it follows that $Y$~is not closed
under~$\Pi$.
\end{proof*}

\begin{lem} \label{tightness}
For any program $\Pi$  and any
consistent set $X$ of literals, if $\Pi$ is tight on $X$ then so is
$\Pi^X$.
\end{lem}

\begin{proof*}
Let $\Pi$ be a program
and $X$ be a consistent set of literals.  Suppose that $\Pi^X$ is not tight on
$X$; we want to show that   $\Pi$ is not tight on $X$.
It is clear from the definition of reduct of a program that
every rule of $\Pi^X$ has the form
\beq
\ii{Head} \lar \ii{Body}^X
\eeq{rule1-r}
for some rule (\ref{rule}) in $\Pi$.
Then there is an infinite sequence $L_1,L_2,\dots$ in $X$ such that,
for every $i$, there is a rule~(\ref{rule1-r}) in $\Pi^X$ with
$L_i = \ii{Head}$,
$L_{i+1} \in \ii{poslit}(\ii{Body}^X) \cap X$ and $X \satisfies \ii{Body}^X$.

By Lemma~\ref{closure-supportedness}(i), $X \satisfies \ii{Body}^X$ iff
\hbox{$X\satisfies \ii{Body}$}.
 From the definition of the reduct,
\hbox{$\ii{poslit}(\ii{Body}^X) \subseteq  \ii{poslit}(\ii{Body})$}.
Therefore,
there is an infinite sequence $L_1,L_2,\dots$ in $X$ such that,
for every $i$, there is a rule~(\ref{rule}) in $\Pi$ with $L_i = \ii{Head}$,
$L_{i+1} \in \ii{poslit}(\ii{Body}) \cap X$ and $X \satisfies \ii{Body}$.
This contradicts the assumption that $\Pi$ is tight on $X$.
\end{proof*}

\begin{proof*}[Proof of Theorem~\ref{fages-extended}
(Section~\ref{sec:thm})]
Consider a program $\Pi$   and a
consistent set $X$ of literals such that $\Pi$ is tight on $X$. Assume
that $X$ is closed under and supported by~$\Pi$. Then, by
Lemma~\ref{closure-supportedness}(ii,iii),  $X$ is closed under and
supported
by~$\Pi^X$.  By Lemma~\ref{tightness}, $\Pi^X$ is tight on $X$.
Hence, by Lemma~\ref{right-to-left}, $X$~is an answer set
for~$\Pi^X$, and consequently an answer set for~$\Pi$.  In the other
direction, the assertion of the theorem follows from
Proposition~\ref{easy-main}.
\end{proof*}

\begin{proof*}[Proof of Proposition~\ref{thm1}
(Section~\ref{sec:tc})]
We will first prove the special case when $\Pi$ doesn't contain
negation as failure.  Let $X$ be an answer set for $\Pi\cup\ii{Def}$;
denote set (\ref{r}) by $R$, and its transitive closure by $R^\infty$.
We need to prove that for all $x$ and $y$, $\ii{tc}(x,y) \in X$  iff
$\langle x,y \rangle \in R^\infty$.

\medskip\noindent{\sl Left-to right.}
Since there is no negation as failure
in $\Pi$, $X$ can be characterized as the union $\bigcup_i X_i$ of the
sequence of sets of literals defined as
follows: $X_0 = \emptyset$; $X_{i+1}$ is the set of all literals $L$
such that $\Pi \cup \ii{Def}$ contains a rule $L \lar \ii{Body}$
with $\ii{Body}$ satisfied by~$X_i$.  We will show by induction on $i$ that
$\ii{tc}(x,y) \in X_i$ implies $\langle x,y \rangle \in R^\infty$.  If $i=0$, the
assertion is trivial because $X_0$ is empty.  Assume that for all
$x$ and $y$, $\ii{tc}(x,y) \in X_i$ implies $\langle x,y \rangle \in R^\infty$,
and take an atom $\ii{tc}(x,y)$ from $X_{i+1}$. Take a rule
$\ii{tc}(x,y) \lar \ii{Body}$ in $\Pi \cup \ii{Def}$ such that
$X_i\satisfies\ii{Body}$. Since $\Pi$
doesn't contain atoms of the form $\ii{tc}(x,y)$ in the heads of rules, this rule
belongs to~\ii{Def}.  {\sl Case~1:} $\ii{Body}=p(x,y)$.  Then
$p(x,y) \in X_i \subseteq X$, so that
$\langle x,y \rangle \in R \subseteq R^\infty$. {\sl Case~2:}
$\ii{Body}=p(x,v),\ii{tc}(v,y)$.  Then $p(x,v) \in X_i \subseteq X$, so that
$\langle x,v \rangle \in R\subseteq R^\infty$; also,
$\ii{tc}(v,y) \in X_i$, so that, by the induction hypothesis,
$\langle v,y \rangle \in R^\infty$. By the transitivity of $R^\infty$, it
follows that $\langle x,y \rangle \in R^\infty$.

\medskip\noindent{\sl Right-to-left.}
Since $R^\infty=\bigcup_{j>0} R^j$, it is sufficient to prove that for all
$j>0$, $\langle x,y \rangle \in R^j$ implies \hbox{$\ii{tc}(x,y) \in X$}.  The proof is by
induction on $j$.  When $j=1$, $\langle x,y \rangle \in R$, so that
$p(x,y) \in X$; since $X$ is closed under \ii{Def}, it follows that
$\ii{tc}(x,y) \in  X$.  Assume that for all $x$ and $y$,
$\langle x,y \rangle \in R^j$ implies $\ii{tc}(x,y) \in X$, and take a pair
$\langle x,y \rangle$ from $R^{j+1}$.  Take $v$ such that
\hbox{$\langle x,v \rangle \in R$} and $\langle v,y \rangle \in R^j$.
Then $p(x,v) \in X$ and, by the induction hypothesis, $\ii{tc}(v,y) \in X$.
Since $X$ is closed under \ii{Def}, it follows that $\ii{tc}(x,y) \in X$.
\end{proof*}

\medskip
We have proved the assertion of Proposition~\ref{thm1}
for programs without negation as
failure.  Now let $\Pi$ be any program that does not contain atoms of the
form $\ii{tc}(x,y)$ in heads of rules, and let $X$ be an answer set for
$\Pi\cup\ii{Def}$.  Clearly, the reduct $\Pi^X$ is a program without negation
as failure that does not contain atoms of the form $\ii{tc}(x,y)$ in the heads of
rules, and $X$ is an answer set for $\Pi^X \cup \ii{Def}$.  By the special
case of the theorem proved above, applied to $\Pi^X$, (\ref{tc}) is the
transitive closure of (\ref{r}).

\begin{proof*}[Proof of Theorem~\ref{thm2}
(Section~\ref{sec:tc})]
Assume (i)--(iii), and assume that $\Pi\cup\ii{Def}$ is not tight on $X$.
Then there exists an infinite sequence $L_1,L_2,\dots$ of elements of $X$
such that for every $i$, $L_{i+1}$  is a parent of $L_i$ relative to
$\Pi\cup\ii{Def}$ and $X$.  Consider two cases.

\medskip\noindent{\sl Case 1:}
Sequence $L_1,L_2,\dots$ contains only a finite number of terms of the form
$\ii{tc}(x,y)$.   Let $L_n$ be the last of them.  Then for every $i>n$,
$L_{i+1}$ is a parent of $L_i$ relative to $\Pi$ and $X$.
Sequence $L_{n+1},L_{n+2},\dots$ shows
that $\Pi$ is not tight on $X$, contrary to (i).

\medskip\noindent{\sl Case 2:}
Sequence $L_1,L_2,\dots$ contains infinitely many terms of the form $\ii{tc}(x,y)$.
By (iii), it follows that this sequence has no terms of the form $p(x,y)$.
The examination of rules \ii{Def} shows that every $\ii{tc}(x,y)$ in
this sequence is immediately followed by a term of the form $\ii{tc}(v,y)$ such that
$p(x,v)\in X$.  Consequently, sequence $L_1,L_2,\dots$ consists of some initial
segment followed by an infinite sequence of literals of the form
$$\ii{tc}(v_0,y),\ii{tc}(v_1,y),\dots$$
such that, for every $i$, $p(v_i,v_{i+1})\in X$.  This is impossible by (ii).
\end{proof*}

\begin{proof*}[Proof of Proposition~\ref{prop1}
(Section~\ref{sec:tc})]
Let $\Pi$ be a program containing constraint (\ref{irref-tc}), with
finite~$C$, and let $X$
be a set of literals closed under $\Pi\cup\ii{Def}$.  Assume that
$\{\langle x,y \rangle:\ p(y,x) \in X\}$ is not well-founded.
Take $x_1,\dots,x_n\in C$ that satisfy~(\ref{finite-seq}) and $x_n=x_1$.
Since $X$ is closed under \ii{Def}, $tc(x_1,x_1)\in X$.  But this is
impossible because $X$ is closed under (\ref{irref-tc}).
\end{proof*}

\begin{proof*}[Proof of Proposition~\ref{prop3}
(Section~\ref{sec:bw})]
Let $\Pi$ be the program that differs from~(\ref{hp1})--(\ref{hp5a}) in that
\begin{itemize}
\item
its underlying set of atoms includes, additionally, expressions of the forms
$\ii{on}(\ii{table},l,t)$ and $\ii{above}(\ii{table},l,t)$, and
\item
rules~(\ref{hp4}) and~(\ref{hp5}) are replaced by
\beq
\ba l
\ii{above}(l,l',t) \lar \ii{on}(l,l',t)\\
\ii{above}(l,l',t) \lar \ii{on}(l,l'',t), \ii{above}(l'',l',t)\\
\ea
\eeq{hp4new}
and
\beq
\lar \ii{above}(l,l,t).
\eeq{hp5new}
\end{itemize}
Let $X$ be a set of literals that does not contain any of the newly introduced
atoms or their negations and is closed under the original
program~(\ref{hp1})--(\ref{hp5a}).  We will prove that $\Pi$ is tight on $X$.
It will follow then that the original program is tight on $X$ as well, because
that program is a subset of $\Pi$.

For every $k=0,\dots,T+1$, let $\Pi_k$ be the subset of the rules of $\Pi$ in
which rules~(\ref{hp4new}) are restricted to $t<k$.  Since $\Pi_{T+1}=\Pi$,
it is sufficient to prove that, for all $k$, $\Pi_k$ is tight on $X$.  The
proof is by induction on $k$.  {\sl Basis:} $k=0$.  The rules of $\Pi_0$ are
(\ref{hp1})--(\ref{hp3a}),~(\ref{hp5a}) and~(\ref{hp5new}).  To see that this
program is tight, define
$$
 \ba l
 \lambda(\ii{on}(l,l',t)) = t + 1, \\
 \lambda(\neg \ii{on}(l,l',t)) = t + 2, \\
 \lambda(\ii{move}(b,l,t)) = \lambda(\neg \ii{move}(b,l,t)) = 0, \\
 \lambda(\ii{above}(l,l',t)) = \lambda(\neg \ii{above}(l,l',t)) = 0.
 \ea
$$
{\sl Induction step:}  Assume that $\Pi_k$ is tight on $X$.  Let $C$ be the
set of location constants, and let functions $p$ and $\ii{tc}$ be defined by
$$
\ba l
p(l,l') = \ii{on}(l,l',k),\\
tc(l,l') = \ii{above}(l,l',k).
\ea
$$
Then $\Pi_{k+1}=\Pi_k\cup\ii{Def}$.  Let us check that all conditions of
Theorem~\ref{thm2} are satisfied.  Condition~(i) holds by the induction
hypothesis.  Since $X$ is closed under the original
program~(\ref{hp1})--(\ref{hp5a}) and does not contain any of the newly
introduced literals, it is closed under $\Pi_{k+1}$ as well; in view of the
fact that $\Pi_{k+1}$ contains constraint~(\ref{hp5new}), condition~(ii)
follows by Proposition~\ref{prop1}.  By inspection,~(iii) holds also.  By
Theorem~\ref{thm2}, it follows that $\Pi_{k+1}$ is tight on~$X$.
\end{proof*}

\section{Conclusion} \label{sec:conclusion}

For absolutely tight logic programs,
the answer set semantics is equivalent to the completion semantics.
Answer sets for a finite normal absolutely tight program can be found by
running a
satisfiability solver on the program's completion.  Defining tightness
relative to a set of literals extends the applicability of this method to
some programs that are not absolutely tight.  This method of computing
answer sets is applicable to rules with nested expressions and with
weight constraints.

Although this method is not directly applicable to
disjunctive programs, disjunction in the head of a rule can be
sometimes eliminated in favor of nested expressions in its
body~\cite{lif99d},~\cite{lif02}.  For instance, the disjunctive rule
$$p; \no\ q \lar r$$
in any program can be replaced by
$$p \lar r, \no\ \no\ q$$
without changing the program's answer sets.

\section*{Acknowledgments}

We are grateful to Yuliya Babovich, Selim Erdo\u gan, Paolo Ferraris,
Joohyung Lee, Victor Marek, Norman McCain
and Emilio Remolina for comments and discussions related to the subject of this
paper.  Special thanks to Hudson Turner for many useful suggestions,
including the use of \ii{poslit} in the definition of tightness
instead of the more complicated condition from~\cite{erd01a}, and the
possibility of restriction~(\ref{b3}) from Section~\ref{sec:tightness}.

This work was partially supported by National
Science Foundation under grant IIS-9732744 and by the Texas
Higher Education Coordinating Board under grant 003658-0322-2001.  The
first author was also
supported by a NATO Science Fellowship.


\begin{thebibliography}{}

\bibitem[\protect\citeauthoryear{Apt, Blair, and Walker}{Apt
  et~al\mbox{.}}{1988}]{apt88}
{\sc Apt, K.}, {\sc Blair, H.}, {\sc and} {\sc Walker, A.} 1988.
\newblock Towards a theory of declarative knowledge.
\newblock In {\em Foundations of Deductive Databases and Logic Programming},
  {J.~Minker}, Ed. Morgan Kaufmann, San Mateo, CA, 89--148.

\bibitem[\protect\citeauthoryear{Babovich, Erdem, and Lifschitz}{Babovich
  et~al\mbox{.}}{2000}]{bab00}
{\sc Babovich, Y.}, {\sc Erdem, E.}, {\sc and} {\sc Lifschitz, V.} 2000.
\newblock Fages' theorem and answer set programming.\footnote{\tt
  http://arxiv.org/abs/cs.ai/0003042 .}
\newblock In {\em Proc.~{NMR}-2000}.

\bibitem[\protect\citeauthoryear{Baral and Gelfond}{Baral and
  Gelfond}{1994}]{bar94}
{\sc Baral, C.} {\sc and} {\sc Gelfond, M.} 1994.
\newblock Logic programming and knowledge representation.
\newblock {\em Journal of Logic Programming\/}~{\em 19,20}, 73--148.

\bibitem[\protect\citeauthoryear{Clark}{Clark}{1978}]{cla78}
{\sc Clark, K.} 1978.
\newblock Negation as failure.
\newblock In {\em Logic and Data Bases}, {H.~Gallaire} {and} {J.~Minker}, Eds.
  Plenum Press, New York, 293--322.

\bibitem[\protect\citeauthoryear{Erdem and Lifschitz}{Erdem and
  Lifschitz}{1999}]{erd99}
{\sc Erdem, E.} {\sc and} {\sc Lifschitz, V.} 1999.
\newblock Transformations of logic programs related to causality and planning.
\newblock In {\em {Logic Programming and Non-monotonic Reasoning: Proc.~Fifth
  Int'l Conf. (Lecture Notes in Artificial Intelligence 1730)}}. 107--116.

\bibitem[\protect\citeauthoryear{Erdem and Lifschitz}{Erdem and
  Lifschitz}{2001a}]{erd01a}
{\sc Erdem, E.} {\sc and} {\sc Lifschitz, V.} 2001a.
\newblock Fages' theorem for programs with nested expressions.
\newblock In {\em Proceedings of the Seventeenth International Conference on
  Logic Programming}. 242--254.

\bibitem[\protect\citeauthoryear{Erdem and Lifschitz}{Erdem and
  Lifschitz}{2001b}]{erd01}
{\sc Erdem, E.} {\sc and} {\sc Lifschitz, V.} 2001b.
\newblock Transitive closure, answer sets, and predicate completion.
\newblock In {\em Working Notes of the {AAAI} Spring Symposium on Answer Set
  Programming}.

\bibitem[\protect\citeauthoryear{Fages}{Fages}{1994}]{fag94}
{\sc Fages, F.} 1994.
\newblock Consistency of {C}lark's completion and existence of stable models.
\newblock {\em Journal of Methods of Logic in Computer Science\/}~{\em 1},
  51--60.

\bibitem[\protect\citeauthoryear{Ferraris and Lifschitz}{Ferraris and
  Lifschitz}{2003}]{fer03}
{\sc Ferraris, P.} {\sc and} {\sc Lifschitz, V.} 2003.
\newblock Weight constraints as nested expressions.
\newblock In progress.

\bibitem[\protect\citeauthoryear{Gelfond and Lifschitz}{Gelfond and
  Lifschitz}{1988}]{gel88}
{\sc Gelfond, M.} {\sc and} {\sc Lifschitz, V.} 1988.
\newblock The stable model semantics for logic programming.
\newblock In {\em {Logic Programming: Proc.~Fifth Int'l Conf.~and Symp.}},
  {R.~Kowalski} {and} {K.~Bowen}, Eds. 1070--1080.

\bibitem[\protect\citeauthoryear{Gelfond and Lifschitz}{Gelfond and
  Lifschitz}{1991}]{gel91b}
{\sc Gelfond, M.} {\sc and} {\sc Lifschitz, V.} 1991.
\newblock Classical negation in logic programs and disjunctive databases.
\newblock {\em New Generation Computing\/}~{\em 9}, 365--385.

\bibitem[\protect\citeauthoryear{Inoue and Sakama}{Inoue and
  Sakama}{1998}]{ino98}
{\sc Inoue, K.} {\sc and} {\sc Sakama, C.} 1998.
\newblock Negation as failure in the head.
\newblock {\em Journal of Logic Programming\/}~{\em 35}, 39--78.

\bibitem[\protect\citeauthoryear{Lifschitz}{Lifschitz}{1996}]{lif96}
{\sc Lifschitz, V.} 1996.
\newblock Two components of an action language.
\newblock In {\em Working Papers of the Third Symposium on Logical
  Formalizations of Commonsense Reasoning}.

\bibitem[\protect\citeauthoryear{Lifschitz}{Lifschitz}{1999}]{lif99c}
{\sc Lifschitz, V.} 1999.
\newblock Answer set planning.
\newblock In {\em Proc.~{ICLP}-99}. 23--37.

\bibitem[\protect\citeauthoryear{Lifschitz}{Lifschitz}{2002}]{lif02}
{\sc Lifschitz, V.} 2002.
\newblock Answer set programming and plan generation.
\newblock {\em Artificial Intelligence\/}~{\em 138}, 39--54.

\bibitem[\protect\citeauthoryear{Lifschitz, Tang, and Turner}{Lifschitz
  et~al\mbox{.}}{1999}]{lif99d}
{\sc Lifschitz, V.}, {\sc Tang, L.~R.}, {\sc and} {\sc Turner, H.} 1999.
\newblock Nested expressions in logic programs.
\newblock {\em Annals of Mathematics and Artificial Intelligence\/}~{\em 25},
  369--389.

\bibitem[\protect\citeauthoryear{Lloyd and Topor}{Lloyd and
  Topor}{1984}]{llo84a}
{\sc Lloyd, J.} {\sc and} {\sc Topor, R.} 1984.
\newblock Making {P}rolog more expressive.
\newblock {\em Journal of Logic Programming\/}~{\em 3}, 225--240.

\bibitem[\protect\citeauthoryear{Simons, Niemel{\"a}, and Soininen}{Simons
  et~al\mbox{.}}{2002}]{sim02}
{\sc Simons, P.}, {\sc Niemel{\"a}, I.}, {\sc and} {\sc Soininen, T.} 2002.
\newblock Extending and implementing the stable model semantics.
\newblock {\em Artificial Intelligence\/}~{\em 138}, 181--234.

\end{thebibliography}
\end{document}